\documentclass{article}

\usepackage{arxiv}
\usepackage{graphicx}
\graphicspath{ {./images/} }

\usepackage{graphics, pdflscape, caption, xcolor}
\usepackage{hyperref, url, cite}
\usepackage{amsmath,amssymb,amsfonts}
\usepackage{algorithmic, textcomp, footnote, booktabs}
\usepackage{orcidlink}

\fancypagestyle{fancyfirst}{
    \fancyhf{}

    \fancyfoot[C]{
        \parbox{\textwidth}{
            \centering
            \scriptsize

            \textbf{Accepted for publication in:}
            \textit{International Conference on Artificial Intelligence, Computer, Data Sciences, and Applications (ACDSA 2027)}\\[2pt]

            \textbf{Research Repository:}
            \href{https://github.com/CarlosP1806/urban-learning-cdmx}
            {Urban Representation Learning for Identifying Potential Disparities in Crime Reporting}\\[4pt]

            \thepage
        }
    }
}

\title{Urban Representation Learning for Identifying Potential Disparities in Crime Reporting}

\author{
 Jorge Valverde-Rebaza\thanks{Corresponding author: jvalverr@tec.mx}\orcidlink{0000-0001-8664-9692} \\
  School of Engineering and Sciences \\
  Tecnologico de Monterrey\\
  Mexico \\
  \texttt{jvalverr@tec.mx} \\
  \And
 Carlos Paez De La Cruz\orcidlink{0009-0004-0459-3509} \\
  School of Engineering and Sciences \\
  Tecnologico de Monterrey\\
  Mexico \\
  \texttt{A017030505@tec.mx} \\
\And
   Pedro Pérez-Collado\orcidlink{0009-0000-5140-314X}\\
  School of Engineering and Sciences \\
  Tecnologico de Monterrey\\
  Mexico \\
  \texttt{plpcollado@tec.mx} \\
\And
 Daniel Sánchez-Cataño\orcidlink{0009-0008-8096-6599}\\
  School of Engineering and Sciences \\
  Tecnologico de Monterrey\\
  Mexico \\
  \texttt{A01663268@tec.mx} \\
\And
 Eashley Martínez-Vergara\orcidlink{0009-0001-7793-5686}\\
  School of Engineering and Sciences \\
  Tecnologico de Monterrey\\
  Mexico \\
  \texttt{A01663894@tec.mx} \\
}

\begin{document}
\maketitle
\thispagestyle{fancyfirst}

\begin{abstract}
Open crime data support urban safety analysis but reflect reported incidents rather than the complete distribution of criminal activity, potentially introducing spatial distortions in downstream analysis. We propose a self-supervised urban representation learning framework for identifying unusual reporting patterns relative to representation-similar peers using only georeferenced crime records. The method constructs complementary hourly, weekly, and static reporting profiles, learns scale-specific representations under geographic self-supervision, and integrates them into a unified latent space. \emph{Neighbor Profile Divergence} (NPD) is then used to quantify disagreement between each region's reporting profile and those of its retrieved peers. Experiments on more than 1.8 million crime records from Mexico City, aggregated into 1,058 H3 regions, show that the proposed representation achieves a mean Silhouette score of 0.4052, compared with 0.2863 for Hex2Vec. NPD rankings remain highly stable across neighborhood sizes, with Spearman correlations from 0.9720 to 0.9951, while most high-divergence candidates persist across configurations. The retrieved peers also extend substantially beyond immediate geographic neighborhoods, indicating that the learned representation captures structure beyond local spatial proximity. High-divergence regions exhibit spatially organized patterns and, in several cases, differences in incident volume relative to peer expectations. These results support NPD as an internal auditing and prioritization mechanism for identifying regions that warrant further investigation, rather than as direct evidence of underreporting.

\keywords{Urban crime analytics \and urban representation learning \and self-supervised learning \and anomaly detection \and crime reporting disparities.}
\end{abstract}


\section{Introduction}
\label{sec:intro}

Urban analytics has become an important paradigm for extracting actionable knowledge from large-scale spatial, temporal, and administrative data. Its applications span mobility modeling, business opportunity identification, smart-city planning, resource allocation, and public safety \cite{Lou:human_mobility:26,valverde:business_op_predict:19,kounadi2020crimeforecasting,deng2023urbanvisual}. At the same time, the increasing availability of open government data has expanded the range of computational methods that can be applied to urban decision-making, enabling analyses that were previously restricted by limited access to administrative information \cite{hagen2019opendata,meijer2019opengovernance}. These developments create opportunities not only to characterize urban dynamics, but also to examine the quality, internal consistency, and limitations of the data on which such analyses depend.

Georeferenced crime records constitute a particularly important example. Police and prosecutorial records provide detailed information about the spatial and temporal distribution of reported events and have been extensively used for crime forecasting, risk estimation, representation learning, and spatial analysis \cite{crivellari2021crimevec,li2022sthypergraph}. Because these records are often available at fine geographic resolution and over long periods, they provide a valuable basis for identifying recurring temporal patterns, spatial concentrations, and differences among urban regions. However, their analytical value must be considered together with an important limitation: administrative crime records do not provide a direct observation of the complete distribution of criminal activity.

For an event to appear in an administrative dataset, it must first become observable to the corresponding institution. Depending on the context, this may require reporting by victims or witnesses, detection by public authorities, institutional access, formal registration, and subsequent classification within an administrative system. Each of these processes may vary geographically and across populations. Consequently, differences in recorded crime can reflect not only differences in underlying criminal activity, but also differences in reporting behavior, institutional practices, policing, and data-recording procedures \cite{builgil2022accuracy,riascos2023underreported,berk2021predictive}. This distinction is particularly relevant for computational analysis because models trained directly on administrative records may reproduce the structure of the reporting process together with the phenomenon they are intended to characterize.

From a methodological perspective, this creates a difficult identification problem. A region with fewer recorded incidents may genuinely experience lower crime, but it may also differ from otherwise comparable regions in its reporting patterns or institutional interaction. Conversely, areas with similar recorded volumes may differ substantially in when incidents are reported, the composition of recorded crime categories, reporting delays, or other structural characteristics. Without independent victimization surveys or institutional indicators, administrative records alone cannot determine which explanation is correct. Nevertheless, they may still contain internal regularities that allow regions with unusual reporting behavior to be identified and prioritized for further investigation.

Urban representation learning offers a natural framework for this type of analysis. Rather than describing each region through isolated variables, representation-learning methods map urban areas into latent spaces in which complex combinations of spatial, temporal, socioeconomic, and behavioral characteristics can be encoded jointly. Such representations have been successfully applied to urban characterization, land-use analysis, socioeconomic prediction, mobility modeling, and crime forecasting \cite{crivellari2021crimevec,li2026multiscale,cao2026urbanmmcl,lin2026mvjc}. By learning similarities directly from high-dimensional regional profiles, these approaches can reveal relationships among areas that may not be evident from geographic proximity or individual indicators alone.

Most existing urban representation-learning methods, however, are designed primarily for prediction or characterization and frequently depend on heterogeneous external information, including mobility traces, points of interest, demographic variables, road networks, or multimodal urban graphs. These sources can improve representation quality, but they also limit applicability in cities where such data are incomplete, unavailable, proprietary, or temporally misaligned. More importantly for the present problem, the resulting representations are rarely used to examine whether the administrative data themselves exhibit unusual internal patterns relative to regions with similar learned characteristics.

This observation motivates a different use of urban representation learning. Instead of treating the learned embedding only as an intermediate representation for a downstream predictive task, we use it to determine which urban regions should be compared. The central intuition is that regions with similar learned representations provide a more informative reference set than regions selected solely by geographic proximity. Once such representation-similar peers are identified, their original reporting profiles can be compared to determine whether a particular region departs substantially from the behavior expected from its peer set. Importantly, this comparison does not assume that representation-similar regions are externally equivalent in socioeconomic or functional terms; rather, similarity is defined strictly by the structure learned from the available crime records.

To operationalize this idea, we develop a self-supervised urban representation-learning framework based exclusively on georeferenced crime records. The framework constructs complementary hourly, weekly, and static reporting profiles, learns view-specific representations using geographic self-supervision, and integrates them into a unified latent space through cross-scale attention. We then introduce \emph{Neighbor Profile Divergence} (NPD), which measures disagreement between the reporting profile of a region and the similarity-weighted profiles of its representation-similar peers. Peer retrieval is performed in the learned embedding space, whereas divergence is evaluated in the reporting-profile space, allowing the framework to distinguish the process used to identify comparable regions from the process used to quantify their disagreement.

NPD is intended as an internal auditing and prioritization mechanism rather than as an estimator of unobserved crime. A high NPD score indicates that a region exhibits a reporting profile that differs from those of regions identified as similar by the learned representation. Such differences may be associated with reporting disparities, but they may also arise from legitimate local heterogeneity, administrative boundaries, land-use differences, policing practices, or other unobserved contextual factors. The proposed framework therefore does not attempt to infer underreporting causally. Instead, it identifies regions whose administrative reporting patterns warrant additional examination using independent evidence.

The main contributions of this work are summarized as follows:

\begin{itemize}
    \item We propose a multi-view self-supervised urban representation-learning strategy that derives regional similarities from a single source of georeferenced crime records, without requiring external mobility, demographic, socioeconomic, or point-of-interest data.

    \item We introduce \emph{Neighbor Profile Divergence} (NPD), an interpretable profile-based measure that quantifies disagreement between a region and its representation-similar peers while separating peer retrieval in the latent space from divergence evaluation in the reporting-profile space.

    \item We evaluate the framework on more than $1.8$ million crime records from Mexico City. The learned representation achieves a mean Silhouette score of $0.4052$, compared with $0.2863$ for Hex2Vec, and the retrieved peer sets extend substantially beyond immediate geographic neighborhoods.

    \item We show that NPD rankings are highly stable across alternative neighborhood sizes, with Spearman correlations ranging from $0.9720$ to $0.9951$, while high-divergence regions exhibit spatially organized patterns and, in several cases, measurable differences between observed and peer-expected incident volumes.
\end{itemize}

Taken together, these contributions position the proposed framework as a method for examining the internal structure of administrative crime records rather than as a replacement for external measurements of crime or victimization. The approach provides a reproducible mechanism for identifying unusual regional reporting patterns and for prioritizing areas where complementary socioeconomic, institutional, or victimization data may be most informative.

The remainder of this paper is organized as follows. Section~\ref{sec:related} reviews related work on urban representation learning and crime reporting disparities. Section~\ref{sec:proposal} presents the proposed framework and the NPD formulation. Section~\ref{sec:experiment} describes the dataset, experimental configuration, baselines, and evaluation protocol. Section~\ref{sec:results} presents the empirical results and discusses their interpretation and limitations. Finally, Section~\ref{sec:conclusion} summarizes the main findings and outlines directions for future research.

\section{Related Work}
\label{sec:related}

\subsection{Urban Representation Learning}

Urban representation learning seeks to encode complex characteristics of urban regions into compact latent representations that preserve relevant spatial, temporal, functional, or behavioral relationships. Rather than describing a region through a collection of isolated indicators, these methods learn multidimensional embeddings in which regions exhibiting similar observed patterns are placed closer together. Such representations can subsequently support a wide range of downstream tasks, including land-use characterization, mobility analysis, socioeconomic inference, recommendation, forecasting, and urban risk assessment.

Early approaches demonstrated that urban functions can be inferred from mobility and geographic interaction patterns, showing that latent representations can capture relationships among regions that are not adequately characterized by physical proximity alone \cite{yao2018urban}. Related work has incorporated geographic context and neighborhood structure directly into the representation-learning objective. Hex2Vec, for example, exploits the spatial organization of H3 cells to learn contextual embeddings of urban areas from geographically structured information \cite{wozniak2021hex2vec}. Geographic representations have also been used to improve urban recommendation and business-opportunity identification, illustrating how spatial context can contribute to the characterization of relationships among urban regions \cite{valverde:business_op_predict:19}.

More recent approaches increasingly combine multiple sources and views of urban activity. Temporal dependencies, graph structure, mobility flows, points of interest, and other heterogeneous signals can be jointly encoded to capture complementary dimensions of regional behavior. Multi-scale temporal graph representation learning has been proposed to model urban dynamics at different temporal granularities \cite{li2026multiscale}, while multimodal and multi-graph contrastive approaches integrate heterogeneous urban information to produce richer regional representations \cite{cao2026urbanmmcl}. Similarly, multi-view learning strategies have shown that combining complementary observations of urban regions can improve the consistency and expressiveness of the resulting embeddings \cite{lin2026mvjc}. Collectively, these studies demonstrate that the notion of similarity among urban regions is inherently multidimensional and can depend on spatial organization, temporal dynamics, mobility, land use, and socioeconomic activity.

Representation learning has also been explored specifically in the context of crime analysis. CrimeVec showed that spatio-temporal embeddings derived from police-recorded crime can capture meaningful relationships among crime categories and urban regions \cite{crivellari2021crimevec}. This line of work is particularly relevant because administrative crime data exhibit strong spatial and temporal regularities and therefore provide a natural source from which regional representations can be learned. However, most existing approaches employ these representations to characterize crime patterns or improve predictive performance. The representation itself is generally treated as an intermediate computational object rather than as a mechanism for examining whether the underlying administrative observations are internally consistent across comparable regions.

A further development is the emergence of generative urban intelligence, in which learned urban representations provide contextual structure for simulation, scenario exploration, and decision-support systems \cite{bittencourt2026generative}. This broader perspective reinforces the importance of reliable urban representations: as embeddings increasingly influence predictive, generative, and decision-support pipelines, understanding what information they encode and what characteristics of the underlying data they reproduce becomes increasingly important.

Despite the progress of urban representation learning, two limitations are particularly relevant to the problem considered in this work. First, many state-of-the-art approaches rely on heterogeneous auxiliary sources, such as mobility traces, points of interest, demographic variables, transportation networks, or complex multimodal graphs. While these sources can enrich the learned representations, they may be unavailable, incomplete, proprietary, or collected at spatial and temporal resolutions that do not align with administrative crime records. This dependence may restrict the applicability and reproducibility of such models across cities.

Second, existing representations are predominantly optimized for prediction, classification, or general urban characterization rather than for auditing the administrative records from which they are learned. Consequently, relatively little attention has been devoted to whether representation learning can provide a principled reference for identifying regions whose observed reporting patterns differ from those of other regions with similar learned characteristics. This distinction is important: our objective is not to infer externally defined urban functions or to claim that nearby points in the embedding space are socioeconomically equivalent. Instead, the learned representation is used to retrieve \emph{representation-similar peers}, which provide an internal reference against which reporting profiles can be compared.

The present work therefore extends urban representation learning in a different direction. We learn multi-view representations directly from georeferenced crime records, without requiring external mobility, demographic, or point-of-interest data, and use the resulting latent space to support peer retrieval rather than a conventional predictive task. The retrieved peers are then used by \emph{Neighbor Profile Divergence} to assess whether a region's reporting profile departs from the behavior observed among regions with similar learned representations. In this way, representation learning becomes part of an internal auditing framework for administrative crime data, complementing its more established use in urban characterization and prediction.


\subsection{Crime Data and Reporting Disparities}

Computational crime analysis commonly relies on police- or prosecution-recorded incidents for forecasting, risk estimation, hotspot identification, and spatial modeling \cite{li2022sthypergraph,riyazuddin2026crimeprediction}. These datasets are attractive because they often provide detailed spatial and temporal information over long periods, allowing computational models to characterize recurring patterns at fine geographic resolutions. However, administrative crime records should not be interpreted as a direct observation of the complete distribution of criminal activity. Rather, they reflect events that become visible to public institutions through reporting, detection, registration, and classification processes.

This distinction is important because each stage of the reporting process may vary across locations and populations. Some incidents may never be reported, while others may be detected through proactive policing or recorded differently depending on institutional procedures. Geographic differences in access to reporting mechanisms, trust in public institutions, policing intensity, administrative practices, and classification procedures may therefore influence the spatial distribution observed in official datasets. As a consequence, variation in recorded crime can reflect a combination of underlying criminal activity and the mechanisms through which that activity becomes administratively observable \cite{builgil2022accuracy,riascos2023underreported,berk2021predictive}.

Previous research has emphasized that these processes can introduce systematic distortions into spatial crime analysis. Underreporting may produce apparent differences between areas that are not fully explained by underlying crime incidence, while heterogeneous recording practices can affect category frequencies, temporal patterns, and geographic concentrations \cite{builgil2022accuracy,riascos2023underreported}. These limitations are particularly relevant for data-driven systems because biases present in the administrative observations can propagate into predictive models, risk scores, and decision-support mechanisms \cite{berk2021predictive}. A model may therefore learn regularities associated not only with crime occurrence, but also with reporting behavior and institutional observation.

Most computational approaches nevertheless treat administrative crime data as the primary analytical input and focus on predicting future events, identifying hotspots, or modeling spatio-temporal dependencies. In such settings, the internal consistency of the reporting process is generally not the primary object of analysis. This creates a methodological gap between using crime records to characterize urban safety and examining whether the records themselves exhibit unusual patterns across otherwise similar regions.

One way to address this gap is to compare a region against a reference set of areas that exhibit similar learned characteristics. If regions are similar in terms of their temporal and compositional reporting structure, substantial deviations in one region may indicate an atypical reporting process that deserves further examination. This idea differs from conventional geographic anomaly detection because the comparison set is not restricted to immediately adjacent areas. Instead, learned representations can provide a data-driven notion of similarity that captures multidimensional relationships among regions.

At the same time, such deviations must be interpreted carefully. Administrative records alone cannot determine whether an unusual profile is caused by underreporting, genuine differences in criminal activity, land use, population dynamics, policing practices, administrative boundaries, or other unobserved contextual factors. Consequently, identifying reporting divergence is not equivalent to estimating missing crime or establishing a causal disparity. Independent information, such as victimization surveys, institutional indicators, or socioeconomic data, would be required to assess the substantive origin of the detected differences.

The present work therefore approaches reporting disparities as an internal consistency problem. Rather than attempting to reconstruct unobserved crime, we ask whether a region's reporting profile is consistent with the profiles observed among its representation-similar peers. This perspective allows administrative data to be examined using their own internal structure while maintaining a clear distinction between anomaly detection and causal inference.

Within this framework, \emph{Neighbor Profile Divergence} provides a mechanism for quantifying departures from peer-expected reporting behavior. The measure is intended to identify regions that warrant additional investigation, not to label them as underreported. In this sense, the proposed approach complements conventional predictive crime analysis by introducing an auditing perspective in which learned urban representations are used to examine the consistency of the administrative records themselves.


\section{Proposal}
\label{sec:proposal}

Fig.~\ref{fig:framework} presents an overview of the proposed framework. It assumes incident-level records containing, at minimum, geographic coordinates, occurrence time, and crime-category information, while optional attributes such as reporting timestamps, offense characteristics, and administrative areas may also be incorporated. Before representation learning, records are cleaned, assigned to spatial units, and aggregated into region-level profiles.

Building on these inputs, the framework learns latent representations of urban regions and uses them to identify reporting patterns that deviate from those of regions with similar learned representations. It comprises three stages: i) construction of complementary regional reporting profiles, ii) self-supervised learning and fusion of scale-specific representations, and iii) retrieval of representation-similar peers followed by \emph{Neighbor Profile Divergence} (NPD) analysis.


\subsection{Multi-Scale Urban Profiles}
\label{subsec:data_representation}

\begin{landscape}

\begin{figure}[p]
    \centering
    \includegraphics[
        width=\linewidth,
        height=0.88\textheight,
        keepaspectratio
    ]{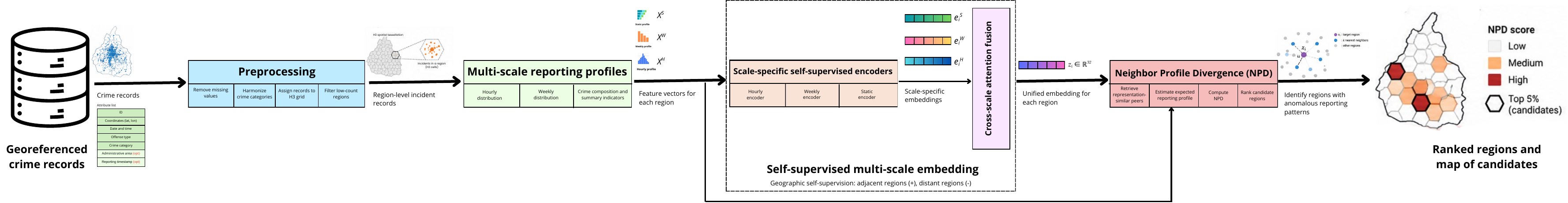}
    \vspace{1.5cm}
    \caption{
    Overview of the proposed urban representation learning framework.
    Georeferenced crime records are first cleaned, harmonized, assigned to H3 spatial units, and aggregated into complementary hourly, weekly, and static reporting profiles.
    Scale-specific encoders are trained through geographic self-supervision and subsequently integrated through cross-scale attention into a unified urban representation.
    For each region, representation-similar peers are retrieved in the learned embedding space, while Neighbor Profile Divergence (NPD) compares their reporting profiles to identify and rank regions exhibiting unusual reporting behavior.
    }
    \label{fig:framework}
\end{figure}

\end{landscape}

Let $\mathcal{H}=\{h_1,\ldots,h_N\}$ denote the spatial units covering the urban area. As shown in the first part of Fig.~\ref{fig:framework}, crime incidents are assigned to their corresponding regions and aggregated into three complementary reporting views.

For each region $h_i$, the hourly profile $\mathbf{x}_i^{H}$ contains the normalized distribution of incidents across the 24 hours of the day. The weekly profile $\mathbf{x}_i^{W}$ combines the day-of-week distribution with the weekday--weekend ratio, temporal entropy, and weekend share. The static profile $\mathbf{x}_i^{S}$ combines crime-category proportions with indicators of crime composition, temporal concentration, reporting delay, and long-term trend. Each view is standardized independently before representation learning.

This separation preserves complementary temporal and compositional structures and allows each view to contribute independently to the learned representation.

\clearpage
\clearpage

\subsection{Self-Supervised Multi-Scale Representation Learning}
\label{subsec:embedding}

Representation learning follows the two-stage process depicted in  Fig.~\ref{fig:framework}. In Stage 1, each reporting view is encoded independently. For each scale $s \in {H,W,S}$, a dedicated encoder $f_{\phi}^{s}$ maps the regional profile into
\begin{equation}
\mathbf{e}_i^{s}
=
f_{\phi}^{s}\left(\mathbf{x}_i^{s}\right).
\label{eq:scale_embedding}
\end{equation}

Geographic contiguity provides the self-supervised signal. Spatially adjacent regions form positive pairs, whereas non-local regions are sampled as negatives. Let $\mathcal{B}$ denote a batch of positive and negative region pairs, with $y_{ij}=1$ for positive pairs and $y_{ij}=0$ otherwise. For each scale, pair similarity is estimated as $\hat{y}_{ij}^{s}=\sigma\left((\mathbf{e}_i^{s})^\top \mathbf{e}_j^{s}\right)$. The three scale-specific binary cross-entropy losses are summed:

\begin{equation}
\begin{aligned}
\mathcal{L}_{P1}
=
-\sum_{s\in\{H,W,S\}}
\frac{1}{|\mathcal{B}|}
\sum_{(i,j)\in\mathcal{B}}
\Big[
y_{ij}\log \hat{y}_{ij}^{s} + (1-y_{ij})\log(1-\hat{y}_{ij}^{s})
\Big].
\end{aligned}
\label{eq:phase1_loss}
\end{equation}

In Stage 2, each scale-specific embedding is compressed by an autoencoder into a common latent space. The three latent vectors are then fused through a single-head cross-scale attention layer and projected into the unified representation $\mathbf{z}_i$. Stage 1 encoders remain fixed, while Stage 2 jointly minimizes reconstruction error and cross-scale contrastive consistency:
\begin{equation}
\mathcal{L}_{P2}
=
\alpha
\sum_{s\in\{H,W,S\}}
\mathcal{L}_{\mathrm{rec}}^{s}
+
(1-\alpha)
\mathcal{L}_{\mathrm{NT\mbox{-}Xent}}.
\label{eq:phase2_loss}
\end{equation}

Here, $\mathcal{L}_{\mathrm{rec}}^{s}$ is the reconstruction loss for scale $s$, $\mathcal{L}_{\mathrm{NT\mbox{-}Xent}}$ promotes cross-view consistency for representations of the same region, and $\alpha$ is learned during training to balance the two objectives.

\subsection{Neighbor Profile Divergence}
\label{subsec:npd}

The final stage of Fig.~\ref{fig:framework} separates two complementary spaces. The learned embedding $\mathbf{z}_i$ determines which regions are compared, whereas reporting profiles quantify how their observed behavior differs. For each region $h_i$, the hourly distribution, day-of-week distribution, and static features are concatenated and normalized into a reporting profile $\mathbf{p}_i$. Let $\mathcal{N}(i)$ denote its $k$ nearest representation-similar peers according to cosine distance in the embedding space.

Let $s_{ij}$ denote the cosine similarity between $\mathbf{z}_i$ and $\mathbf{z}_j$, and define $\tilde{s}_{ij}=\max(s_{ij},0)$. The similarity-based peer weight and expected reporting profile are
\begin{equation}
w_{ij}
=
\frac{\tilde{s}_{ij}}
{\sum_{\ell\in\mathcal{N}(i)}
\tilde{s}_{i\ell}+\varepsilon},
\qquad
\bar{\mathbf{p}}_i
=
\sum_{j \in \mathcal{N}(i)}
w_{ij}\mathbf{p}_j,
\label{eq:expected_profile}
\end{equation}

where $\varepsilon>0$ provides numerical stability. Since each $\mathbf{p}_i$ is normalized to unit length but the peer-weighted profile $\bar{\mathbf{p}}_i$ is not renormalized, we define \emph{Neighbor Profile Divergence} as
\begin{equation}
\mathrm{NPD}(i)
= 1-
\mathbf{p}_i^{\top}\bar{\mathbf{p}}_i.
\label{eq:npd}
\end{equation}

Higher NPD values indicate greater disagreement between a region's reporting profile and the profiles of its representation-similar peers. Because the individual profiles are unit-normalized, NPD reflects the weighted disagreement between a region and its representation-similar peers. Higher values therefore indicate lower weighted profile similarity to the retrieved peer set.

As a complementary volumetric indicator, we define the peer-expected incident count as
$\bar{n}_i=\sum_{j\in\mathcal{N}(i)}w_{ij}n_j$
and the stabilized count ratio as
$\rho_i=n_i/(\bar{n}_i+1)$.

NPD separates peer identification from divergence evaluation: peers are retrieved in the learned representation space, whereas divergence is measured in the reporting-profile space. This differs from anomaly scores based directly on latent distance, reconstruction error, geographic adjacency, or local density, and allows reporting inconsistencies to be examined independently from spatial proximity or raw incident volume.

NPD should nevertheless be interpreted as a screening and auditing measure rather than as an estimator of unobserved crime. High divergence does not establish underreporting causally, since local heterogeneity, administrative practices, or other unobserved contextual factors may produce similar patterns. The resulting ranking therefore identifies candidate regions whose reporting behavior warrants further investigation.

\section{Experimental Setup}
\label{sec:experiment}

We evaluate representation quality and NPD robustness using the dataset, model configuration, baselines, and evaluation protocol described below.

\subsection{Dataset and Preprocessing}
\label{subsec:dataset}

We use open georeferenced crime records published by Mexico City's \emph{Fiscalía General de Justicia} (FGJ). The dataset contains $2,098,743$ records with report dates spanning 2016 to January 2025 and includes spatial, temporal, crime-category, offense, and administrative information. Records with missing spatial or temporal attributes or offense labels not covered by the harmonized crime taxonomy are removed, yielding $1,862,927$ valid records.

Following Fig.~\ref{fig:framework}, records are aggregated using the H3 grid at resolution 8. Cells with fewer than $30$ records are discarded to reduce instability in sparse profiles, yielding $1,058$ regions. The hourly, weekly, and static views defined in Section~\ref{subsec:data_representation} are then constructed and standardized independently.


\subsection{Model Configuration}
\label{subsec:model_setup}

Stage 1 uses separate ReLU encoders for the hourly, weekly, and static views, with dropout $0.2$, a hidden width of $32$, and output dimensions $32$, $8$, and $64$, respectively. Adjacent H3 regions form positive pairs, while three negative pairs per anchor are sampled outside the second H3 neighborhood ring. Training uses Adam with learning rate $6.626250\times10^{-4}$, batch size $64$, and $20$ epochs.

In Stage 2, each scale-specific embedding is compressed into an 8-dimensional latent representation and the three latents are fused through single-head attention into a 32-dimensional representation $\mathbf{z}_i$. Stage 1 encoders remain fixed. Training uses Adam with learning rate $2.075641\times10^{-3}$, batch size $256$, and $50$ epochs. The NT-Xent temperature is $0.07$, and the learned weighting parameter converges to $\alpha=0.708$. Both stages use default PyTorch initialization and seed $32$, without early stopping, learning-rate scheduling, or weight decay.

For NPD, the 24-dimensional hourly distribution and the seven day-of-week components are normalized independently, while the 45 static features are scaled to $[0,1]$. These components are concatenated into a 76-dimensional profile and normalized to unit length. Representation-similar peers are retrieved by cosine distance in the learned embedding space. We use $k=32$, $\varepsilon=10^{-9}$ for numerical stability, and designate the highest 5\% of NPD scores as candidates for further investigation. Sensitivity is assessed with $k \in \{10,16,22,32,45\}$.

\subsection{Baselines and Evaluation Protocol}
\label{subsec:baselines}

We compare the proposed model against PCA, NMF, a dense autoencoder, and Hex2Vec \cite{wozniak2021hex2vec}, covering linear, nonlinear, and spatial representation strategies. All baselines use the same 79-dimensional concatenated reporting profile and produce 32-dimensional embeddings. Hex2Vec additionally uses the same geographic positive and negative pairs adopted in Stage 1. Hyperparameter searches for the proposed model, autoencoder, and Hex2Vec use the same Optuna objective and trial budget, while PCA and NMF require no comparable capacity tuning once the output dimensionality is fixed.

Because external functional labels are unavailable, representation quality is evaluated through $k$-means clustering over $K \in \{6,12,17,23,34,46,69\}$. We report the Silhouette Score, Calinski--Harabasz Index, and Davies--Bouldin Index to assess embedding structure across different cluster granularities. NPD robustness is evaluated separately by comparing rankings obtained for alternative neighborhood sizes against the reference configuration $k=32$, using Spearman rank correlation and overlap among the top 5\% highest-divergence regions.

\section{Results and Discussion}
\label{sec:results}

We assess the structural quality of the learned representation and the stability and spatial interpretability of NPD. The results are treated as evidence of unusual reporting patterns relative to representation-similar peers, rather than as direct evidence of underreporting.

\subsection{Quality of the Learned Urban Representation}
\label{subsec:representation_results}

The proposed anomaly analysis requires a structured embedding in which nearest-neighbor comparisons are meaningful. Because no external functional labels are available, clustering metrics are interpreted as measures of internal organization rather than as evidence of functional equivalence. We therefore compare the learned representation against PCA, NMF, a dense autoencoder, and Hex2Vec.

\begin{figure}[h]
    \centering
    \includegraphics[width=\columnwidth]{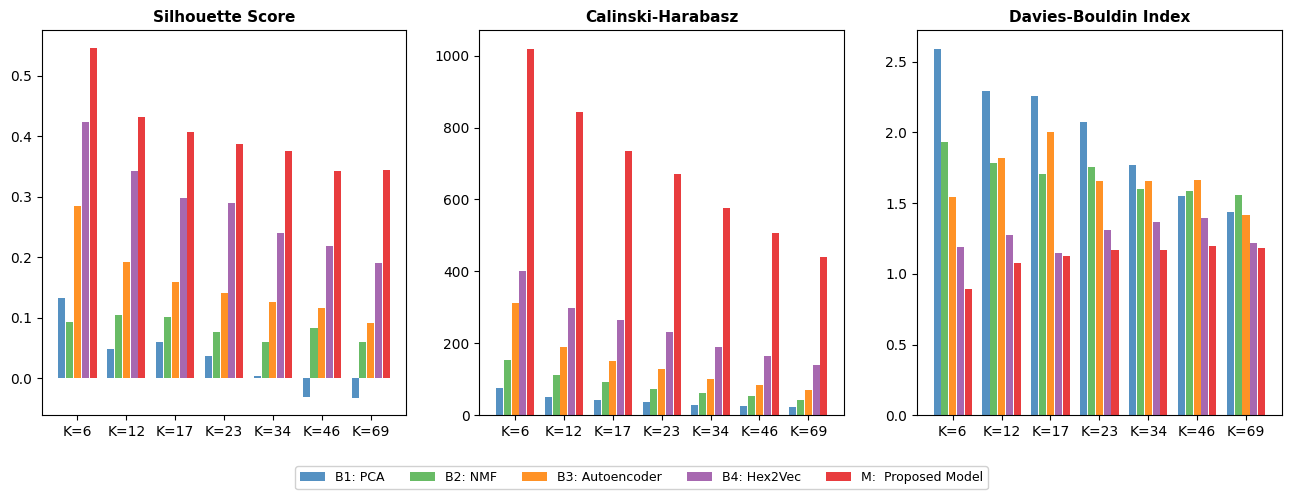}
    \caption{Clustering quality across different values of $K$. Higher Silhouette and Calinski--Harabasz scores and lower Davies--Bouldin values indicate better structured embedding spaces.}
    \label{fig:embedding_quality}
\end{figure}

Figure~\ref{fig:embedding_quality} reports Silhouette, Calinski--Harabasz, and Davies--Bouldin scores across the evaluated values of $K$. The proposed model achieves the best clustering structure at every configuration. Its mean Silhouette score is $0.4052$, compared with $0.2863$ for Hex2Vec, the strongest baseline, representing an improvement of approximately $42\%$. The mean Calinski--Harabasz score increases from $241.7$ for Hex2Vec to $684.4$, while the Davies--Bouldin score decreases from $1.2726$ to $1.1149$, consistently indicating a more structured latent space.

The baseline comparison also clarifies the contribution of spatial and multi-view information. PCA and NMF capture global structure without nonlinear or geographic supervision, the autoencoder introduces nonlinear compression, and Hex2Vec incorporates geographic contiguity. Its stronger performance relative to the other baselines is consistent with the relevance of spatial context, while the additional improvement achieved by the proposed model is consistent with the benefit of combining geographic self-supervision with complementary reporting views.

Importantly, representation-similar peers are not merely local geographic neighbors. Only $7.0\%$ of the retrieved peers lie within two H3 rings, compared with $48.2\%$ for the geographically nearest regions, and their median spatial distance is $10.7$ km versus $2.6$ km for geographic neighbors. The mean overlap between both peer sets is $11.7\%$. These results indicate that the learned embedding retains spatial structure without reducing peer retrieval to geographic proximity.

These results do not directly validate NPD, but they establish a necessary basis for its interpretation. Since NPD compares regions selected in the learned embedding space, stronger internal organization and nontrivial peer structure reduce the likelihood that subsequent comparisons are driven solely by arbitrary or purely local geographic relationships.

\subsection{NPD Stability and Spatial Distribution}
\label{subsec:npd_results}

We next examine whether the regions prioritized by NPD depend strongly on the number of representation-similar peers. NPD is recomputed for $k\in{10,16,22,32,45}$ and compared with the reference configuration $k=32$. Spearman correlations between complete rankings range from $0.9720$ to $0.9951$, while overlap among the top $5\%$ highest-divergence regions ranges from $0.8846$ to $1.0000$. Moreover, $45$ of the $53$ regions identified at $k=32$ remain in the top $5\%$ for all evaluated neighborhood sizes. These results indicate that both the overall ranking and the highest-priority candidates are largely stable with respect to $k$.

\begin{figure}[t]
    \centering
    \includegraphics[width=\columnwidth]{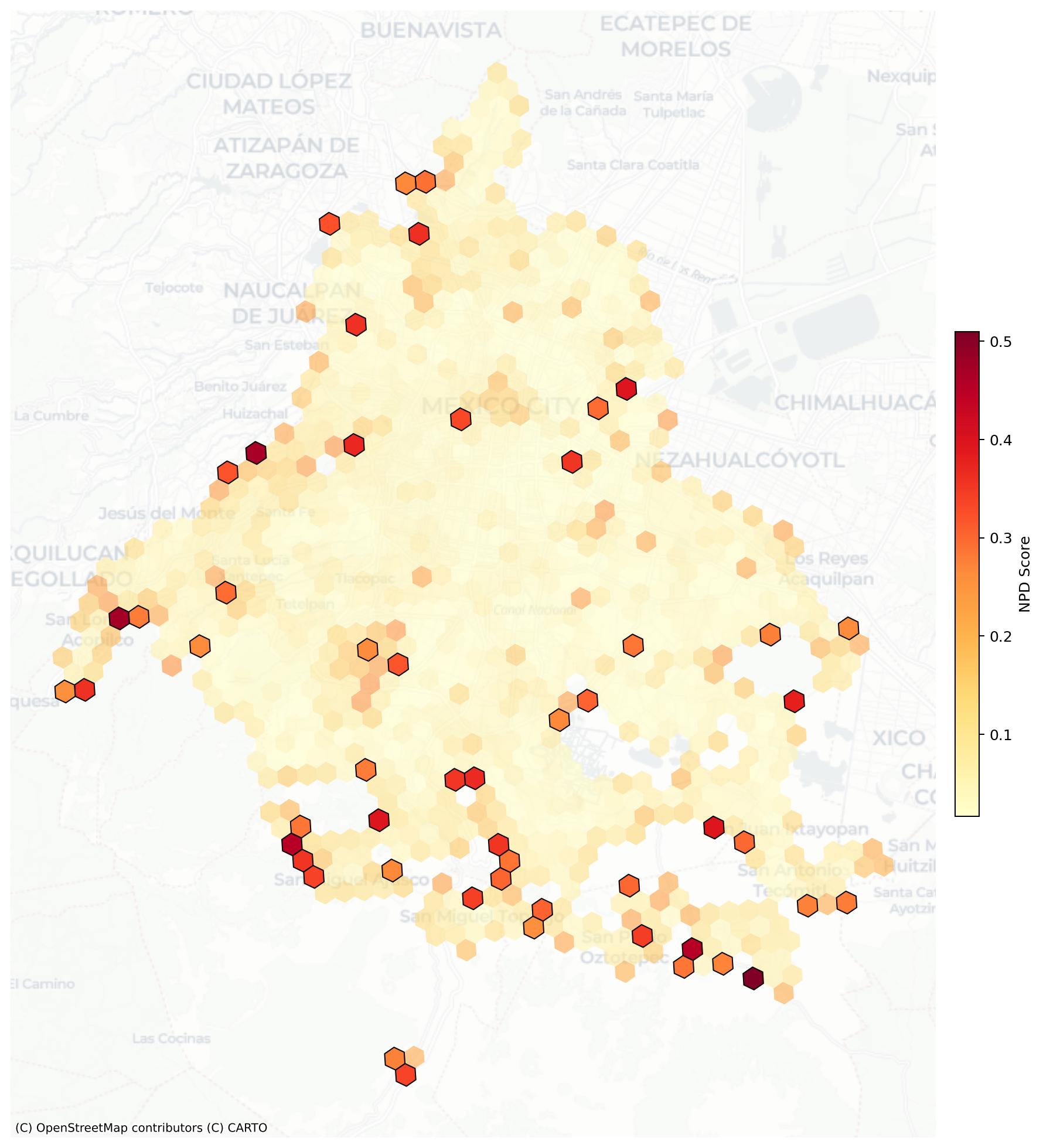}
    \caption{Spatial distribution of Neighbor Profile Divergence across Mexico City. Regions in the highest $5\%$ of NPD values are highlighted as candidates for further investigation.}
    \label{fig:npd_map}
\end{figure}

We therefore use $k=32$ to examine the spatial distribution of reporting divergence. Figure~\ref{fig:npd_map} maps NPD across Mexico City and highlights the highest $5\%$ of scores. Candidate regions exhibit a non-uniform spatial distribution, with several appearing in southern and peripheral areas, including Milpa Alta, Tlalpan, Xochimilco, and Tláhuac, while others occur in more central or western alcaldías.

Table~\ref{tab:top_npd} reports the five highest-NPD regions together with their observed and peer-expected incident counts. All five exhibit fewer observed incidents than expected from their representation-similar peers, with stabilized count ratios ranging from $0.441$ to $0.799$.

\begin{table*}[t]
\centering
\caption{Top-ranked regions according to Neighbor Profile Divergence (NPD). 
Observed incident counts are compared with similarity-weighted peer expectations, 
highlighting regions whose reporting profiles exhibit the strongest deviations from 
their representation-similar peers.}
\label{tab:top_npd}

\small
\setlength{\tabcolsep}{10pt}
\renewcommand{\arraystretch}{1.20}

\begin{tabular}{clrrrr}
\toprule
\textbf{Rank} &
\textbf{Alcaldía} &
\textbf{NPD} &
\textbf{Observed} &
\textbf{Peer Expected} &
\textbf{Count Ratio} \\
\midrule
1 & Milpa Alta             & \textbf{0.510} & 33 & 60 & 0.538 \\
2 & Cuajimalpa de Morelos  & 0.473 & 45 & 65 & 0.680 \\
3 & Miguel Hidalgo         & 0.468 & 36 & 80 & 0.441 \\
4 & Tlalpan                & 0.455 & 32 & 52 & 0.602 \\
5 & Milpa Alta             & 0.454 & 43 & 52 & 0.799 \\
\bottomrule
\end{tabular}

\end{table*}

These cases also illustrate that NPD is not simply a low-count detector. The score is derived from disagreement between normalized reporting profiles and representation-similar peers, whereas incident volume is evaluated separately through the count ratio. The low ratios observed among the highest-ranked cases therefore provide complementary evidence that profile divergence can coincide with differences in reporting volume without being defined by them.

The spatial concentration of some candidates should nevertheless be interpreted cautiously. Peripheral location does not establish weaker reporting or institutional access, and the available data provide no independent ground truth for unreported crime. The identified regions should therefore be treated as priorities for subsequent investigation using independent socioeconomic, victimization, or institutional-access data.


\subsection{Relationship Between Profile and Volumetric Divergence}
\label{subsec:volume_results}

To examine whether NPD captures information beyond incident volume, we compare it with the absolute deviation of each region's stabilized count ratio from parity. Because incident counts are not used directly to compute NPD, this comparison provides an internal consistency check between profile-based and volumetric discrepancies.

\begin{figure}[t]
    \centering
    \includegraphics[width=\columnwidth]{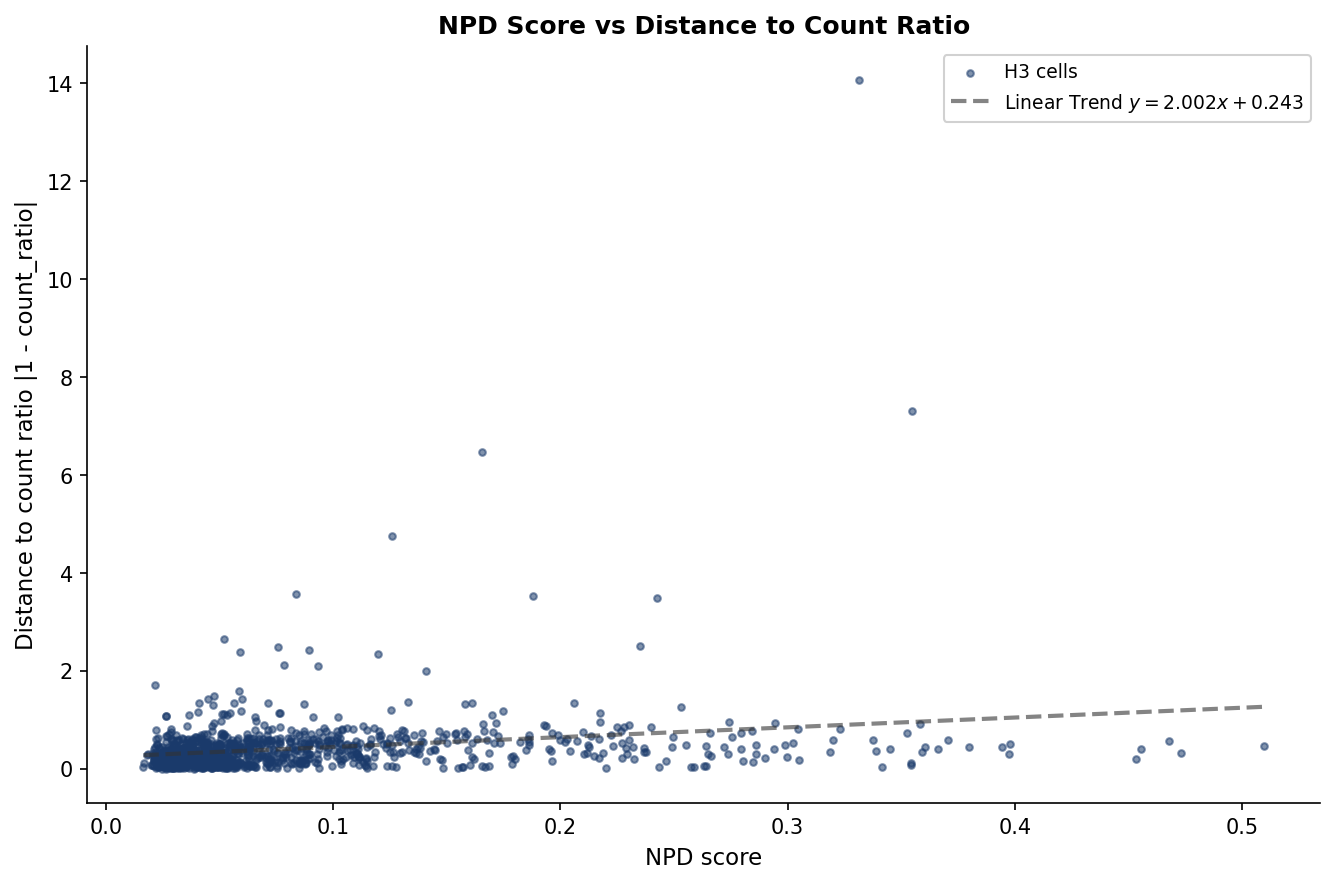}
    \caption{Relationship between NPD and deviation from incident-count parity. Higher profile divergence tends to coincide with larger differences between observed and peer-expected incident volumes.}
    \label{fig:npd_count}
\end{figure}

Figure~\ref{fig:npd_count} shows a positive association between NPD and count-ratio deviation. Spearman correlation indicates a moderate but statistically significant relationship, with $\rho=0.358$ and $p<10^{-6}$.

The moderate correlation indicates that profile and volumetric divergence are related but distinct. A substantially stronger association would suggest that NPD largely reproduces differences in incident volume, whereas the observed relationship shows that regions with unusual reporting profiles do not necessarily exhibit proportional count discrepancies. NPD should therefore be interpreted as a profile-based measure of reporting inconsistency rather than as a proxy for low incident counts.

Taken together, the experiments show that the proposed framework produces a structured representation that supports stable comparison among representation-similar regions. NPD rankings remain robust across neighborhood sizes, while high-divergence regions exhibit spatial organization and, in several cases, measurable differences in incident volume. These results support the use of georeferenced crime records to prioritize regions with unusual reporting behavior for further investigation.

The evidence nevertheless does not support causal attribution. NPD cannot distinguish potential reporting disparities from local heterogeneity, administrative boundaries, policing practices, or other unobserved contextual factors. Moreover, systematic differences shared across representation-similar regions may remain difficult to identify because their reporting profiles can remain mutually consistent. External data, such as victimization surveys or institutional indicators, are therefore required to determine whether individual candidates correspond to actual underreporting. Within these limits, the framework should be interpreted as an auditing and prioritization mechanism.

\section{Conclusion}
\label{sec:conclusion}

This work presented a self-supervised urban representation learning framework for identifying unusual patterns in georeferenced crime records that may indicate potential disparities in crime reporting. By integrating hourly, weekly, and static reporting views, the framework learns a unified latent representation for retrieving representation-similar peers and introduces \emph{Neighbor Profile Divergence} (NPD) to quantify disagreement between a region's reporting profile and those of its peer set.

The learned representation achieves stronger internal organization than the evaluated linear, nonlinear, and spatial baselines, with a mean Silhouette score of $0.4052$ compared with $0.2863$ for Hex2Vec. NPD rankings remain highly stable across neighborhood sizes, with Spearman correlations from $0.9720$ to $0.9951$, and high-divergence regions exhibit spatially organized patterns and, in several cases, differences between observed and peer-expected incident volumes. These results support the use of the framework as a reproducible mechanism for prioritizing regions with unusual reporting behavior.

These findings should nevertheless be interpreted cautiously. Representation learning and NPD rely on related reporting information, and an unusual profile cannot by itself establish underreporting or any other causal mechanism. Local heterogeneity, land use, population dynamics, policing practices, administrative boundaries, and institutional factors may produce similar deviations. NPD should therefore be understood as an internal auditing and prioritization mechanism rather than as direct evidence of unobserved crime.

The framework can be adapted to other cities with sufficiently detailed georeferenced administrative crime data. Such applications should rely on local retraining rather than direct transfer of the Mexico City representation, since crime taxonomies, reporting procedures, spatial density, and institutional practices may differ substantially across jurisdictions. The same pipeline can nevertheless be preserved: construction of multi-view regional profiles, geographically self-supervised representation learning, retrieval of representation-similar peers, and NPD-based prioritization. Future work will investigate cross-city adaptation, robustness across spatial and temporal resolutions, and external validation using victimization surveys and institutional indicators. The implementation and supporting material are available at \url{https://github.com/CarlosP1806/urban-learning-cdmx}.

\section*{Acknowledgments}

The authors gratefully acknowledge the Department of Computer Science, Mexico City Campus, and FAP at Tecnologico de Monterrey for their support of this research.

\bibliographystyle{IEEEtran}
\bibliography{references}

\end{document}